\documentclass[letterpaper, 10 pt, conference]{ieeeconf}  
\usepackage[T1]{fontenc}
\usepackage{graphicx}
\usepackage{amsfonts}
\usepackage{amssymb}
\usepackage{amsmath}
\usepackage{listings}
\usepackage{dirtytalk}
\usepackage{xcolor}
\usepackage{multirow}
\usepackage{hyperref}

\hypersetup{
    colorlinks=true,
    linkcolor=blue,
    urlcolor=blue,
    pdftitle={Overleaf Example},
    pdfpagemode=FullScreen,
}

\IEEEoverridecommandlockouts                              

\title{\LARGE \bf
    Making the Flow Glow -- Robot Perception under Severe Lighting Conditions using Normalizing Flow Gradients
}

\author{Simon Kristoffersson Lind$^{1,*}$ \and Rudolph Triebel$^{2}$ \and Volker Kr\"uger$^{1}$
\thanks{$^{1}$Lund University LTH, Lund, Sweden
        {\tt\small simon.kristoffersson\_lind@cs.lth.se, volker.krueger@cs.lth.se}}
\thanks{$^{2}$German Aerospace Center, Oberpfaffenhofer, Germany,
        and Karlsruhe Institute of Technology, Karlsruhe, Germany
        {\tt\small rudolph.triebel@dlr.de}}%
\thanks{$^{*}$Corresponding Author}%
\thanks{
    \color{gray}
    This is a Preprint, and aside from formatting differences it is identical to the version accepted for publication at IEEE IROS 2024.
}
}

\newcommand{\x}{\mathbf{x}}
\renewcommand{\a}{\pmb{\alpha}}
\renewcommand{\b}{\pmb{\beta}}
\renewcommand{\u}{\mathbf{u}}
\newcommand{\N}{\mathcal{N}}

\begin{document}

\maketitle
\thispagestyle{empty}
\pagestyle{empty}

\begin{abstract}

Modern robotic perception is highly dependent on neural networks.
It is well known that neural network-based perception can be unreliable in real-world deployment,
especially in difficult imaging conditions.
Out-of-distribution detection is commonly proposed as a solution for ensuring reliability in real-world deployment.
Previous work has shown that normalizing flow models can be used for out-of-distribution detection to
improve reliability of robotic perception tasks.
Specifically, camera parameters can be optimized with respect to the likelihood output from a normalizing flow,
which allows a perception system to adapt to difficult vision scenarios.
With this work we propose to use the absolute gradient values from a normalizing flow,
which allows the perception system to optimize local regions rather than the whole image.
By setting up a table top picking experiment with exceptionally difficult lighting conditions,
we show that our method achieves a 60\% higher success rate for an object detection task
compared to previous methods.

\end{abstract}

\section{INTRODUCTION}
In robotics, neural networks have emerged as an indispensable tool for perception and computer vision tasks.
However, they operate under the assumption that all inputs come from the same distribution as their training data.
In other words, when confronted with unfamiliar inputs, their outputs can be unpredictable \cite{OODSurvey}.
This unpredictability poses a significant risk for robots operating in dynamic real-world settings.
Moreover, the accuracy of computer vision hinges on the camera's ability to produce clear images for the neural network.
A case in point is the 2015 Tesla crash,
where the autopilot system failed to detect a truck's side against a bright sky \cite{teslacrash}.

One common way to prevent unpredictable behaviour is out-of-distribution (OOD) detection \cite{OODSurvey}.
Most OOD methods work by directly or indirectly approximating the probability density function (PDF) of the training data.
The approximate PDF is then used as a score function to determine whether an input is in- or out-of-distribution.
Using such a score, an autonomous system can react by performing a safety behaviour
whenever an input is deemed too far from the training distribution.
Most commonly, this safety behaviour simply discards any OOD input while waiting for better inputs.

\begin{figure}[h!]
    \vspace{1em}
    \centering
    \parbox{.45\textwidth}{
        \centering
        \includegraphics[width=.45\textwidth]{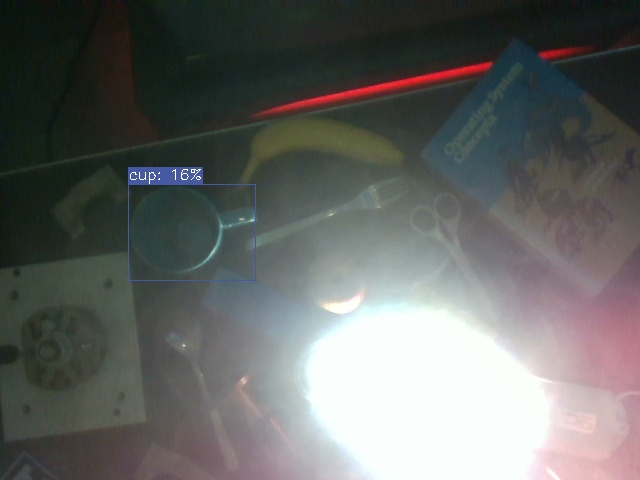} \\
        \includegraphics[width=.45\textwidth]{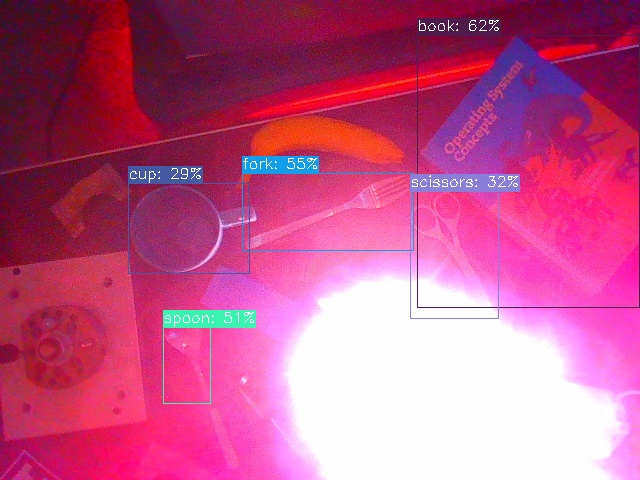}
    }
    \caption{
        In the top image, taken with default camera parameters and auto-exposure,
        only the cup can be detected by YOLOv4.
        In the bottom image, where camera parameters have been optimized using our proposed method,
        the cup, spoon, fork, scissors, and book are successfully detected.
    }
    \label{fig:initial_example}
\end{figure}
\begin{figure}[h!]
    \vspace{1em}
    \centering
    \centering
    \parbox{.45\textwidth}{
        \centering
        \includegraphics[width=.45\textwidth]{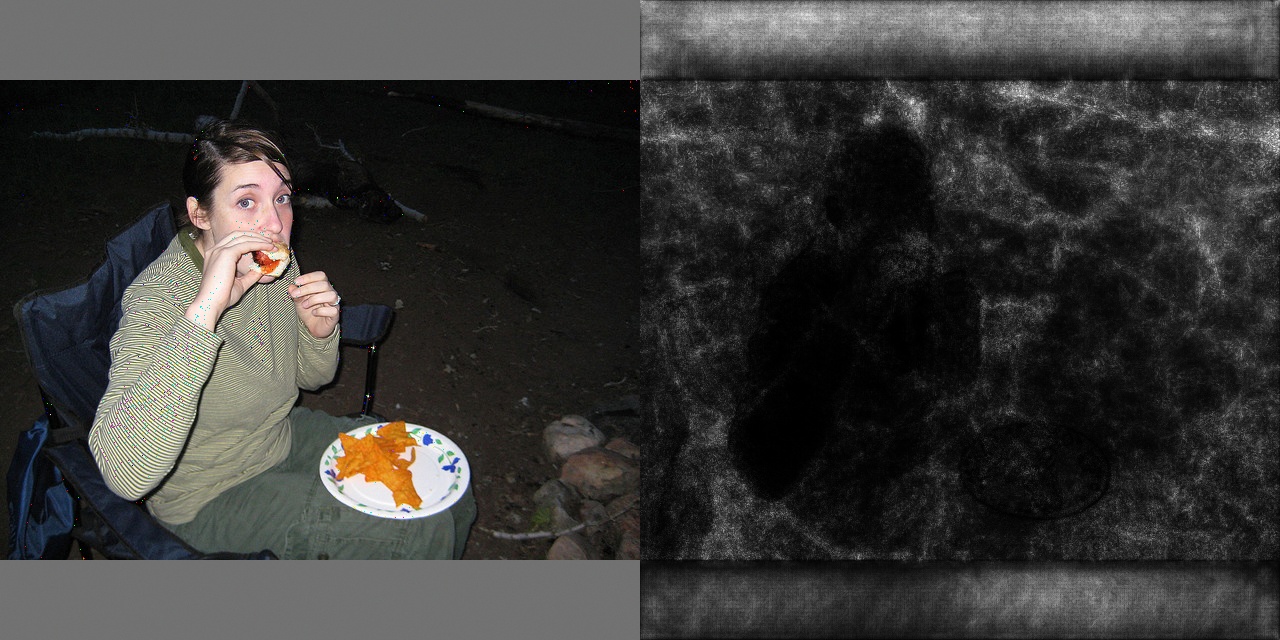} \\
        \includegraphics[width=.45\textwidth]{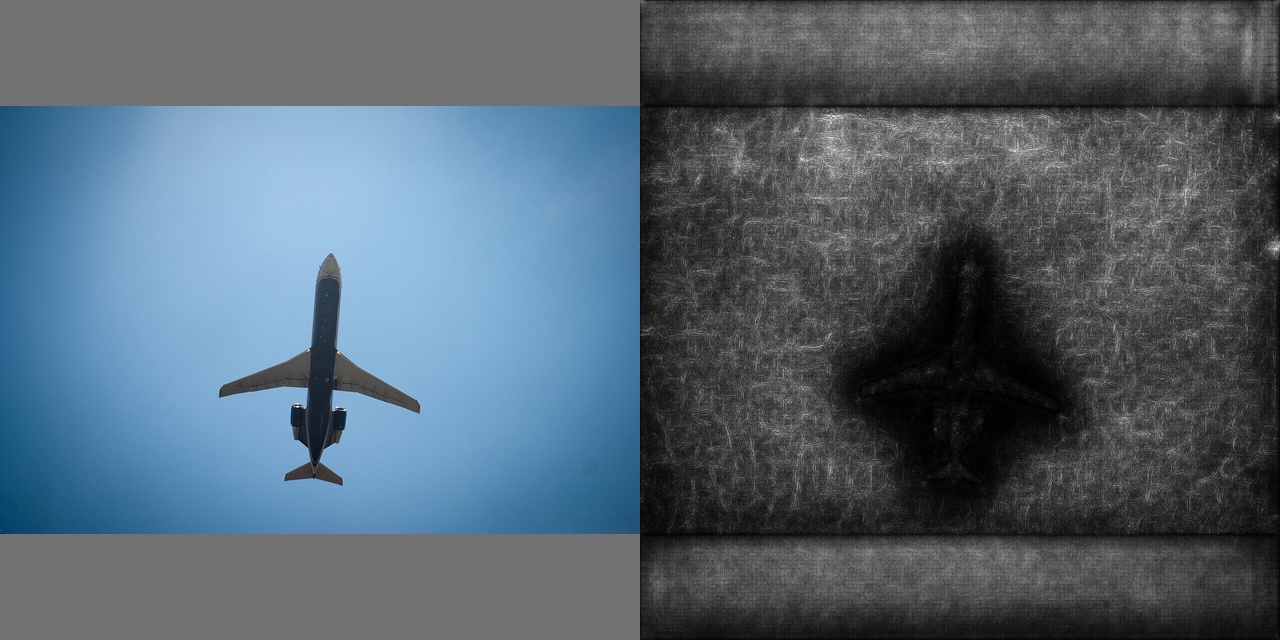}
    }
    \caption{
        Left: Images from the COCO validation set. \\
        right: Absolute gradient images from YOLOv4 + Normalizing Flow.
    }
    \label{fig:yolo_qual}
\end{figure}

While discarding inputs may be feasible for many applications,
it can be problematic for many autonomous systems.
Consider an autonomous car that encounters OOD data.
Should it keep driving while discarding inputs, and risk hitting someone?
Or should it stop while waiting for better inputs, and risk an accident by blocking traffic?
We argue that it is necessary for autonomous systems to detect and adapt to OOD data.

In our earlier work \cite{scia23},
we used a Normalizing Flow (NF) as an OOD detector,
and we created an adaptable vision system by allowing our robot to change parameters in its camera on-the-fly.
We chose to implement this behaviour as an optimization procedure,
where the parameters \textit{backlight compensation, brightness, exposure, gain, saturation, and sharpness}
were optimized with respect to the OOD score from our NF.
Effectively, our robot learns the best camera parameters for any given scene, by minimizing the OOD score.
We were able to show that this type of adaptive behaviour improves the performance of a
YOLOv4 object detector in difficult visual scenarios.

A major weakness of our earlier method is the fact that the NF evaluates an entire image as either in- or out-of-distribution.
Consider the example in Fig. \ref{fig:initial_example}.
Due to the bright light, it is not possible to find any single camera parameter setting that makes the entire image visible.
Despite this, the objects around the bright light can be made almost perfectly visible by changing camera parameters.
Therefore, we see a need to express OOD detection on regions rather than the full image.

When inspecting the pixel-wise gradient with regard to the OOD score from our NF model,
it seems that OOD regions correspond to large absolute gradient values,
and in-distribution regions correspond to small absolute gradient values.
This effect can be seen in Fig. \ref{fig:yolo_qual}.

In this work, we propose to use the absolute pixel-wise NF gradient values as a local OOD score.
We conduct two experiments to provide empirical evidence that the absolute NF gradient values carry OOD information at the pixel-level.
Finally, we show that our proposed absolute NF gradient outperforms our previous method \cite{scia23} on a real-world
robotic picking experiment in a difficult visual scenario.
To summarize, we make the following contributions:
\begin{itemize}
    \item We propose to use the absolute gradient from an NF to provide pixel-level OOD information.
    \item We empirically show the efficacy of our proposed NF gradient for OOD detection.
    \item We show, across two common object detection architectures
          (YOLOv4\cite{YOLOv4}, Faster-RCNN\cite{FasterRCNN})
          that object detectors tend to perform better in regions with small absolute NF gradient values.
    \item We experimentally establish the superior performance of our method over \cite{scia23}
          in adapting camera parameters for challenging imaging scenarios.
\end{itemize}

\section{Related Work}
A substantial body of research focuses on OOD detection, such as \cite{Deecke2019,Hsu2020,Liang2017,Winkens2020,Ren2019,Mohseni2020}.
However, these studies predominantly validate their findings using benchmarks.
The benchmarks employed generally fail to capture the diverse scenarios a robot might face in real-world operations.
This limitation arises primarily because these works ensure that in-distribution and out-of-distribution inputs come from distinct sets.

A more niche segment of research delves into the application of OOD detection in robotics, for example \cite{yuhas2021,mcallister2019}.
These studies, while applying OOD detection to robotics, typically resort to aborting operations upon encountering OOD data.

Another research avenue explores OOD techniques for visual anomaly detection, with studies like \cite{du2020,samuel2021,salimpour2022}.
These investigations typically employ autoencoder methods to identify OOD elements within images, such as in medical imaging contexts.
However, the OOD strategies in these studies are usually tailored to their specific applications, limiting their broader applicability.

Lastly, \cite{scia23} leverages an NF to adapt to challenging imaging situations by fine-tuning camera parameters. Our contribution enhances this by using the NF's gradient to incorporating pixel-level OOD information.

\section{Background}
Given a training distribution $p_{\text{train}}(\x)$, $\x \in \mathbb{R}^D$,
the goal of OOD detection is to identify whether or not a new sample $\hat \x$ comes from $p_{\text{train}}(\x)$.
We assume that $p_{\text{train}}(\x)$ is unknown or intractable to evaluate -- otherwise OOD detection would be trivial.
NFs enable OOD detection by approximating $p_{\text{train}}$ \cite{kirichenko}.

Formally, an NF is a \textit{diffeomorphism},
an invertible transformation $T(\x)$ where both $T(\x)$ and $T^{-1}(\x)$ are differentiable \cite{papamakarios}.
In training an NF, the goal is to learn this diffeomorphism such that
$\u = T(\x)$, $\u \in \mathbb{R}^D$, follows a different, tractable distribution $p_u(\u)$.
The distribution $p_u(\u)$ is typically chosen to be a unit Gaussian $\N(\mathbf{0}, \mathbf{1})$,
since a Gaussian is easy to evaluate and sample.
$T$ being a diffeomorphism allows $p_{\text{train}}(\x)$ to be approximated using the following formula \cite{papamakarios}:
\begin{align*}
    p_{\text{train}}(\x) &= p_u(\u) | \det J_{T^{-1}}(\u) |^{-1} \\
    &= p_u(T(\x)) | \det J_T(\x) | \enskip,
\end{align*}
where $J_T$ denotes the Jacobian of $T$.
As such, the NF allows us to approximate $p_{\text{train}}$ by transforming it into a tractable distribution.
In the standard OOD detection setting, an OOD threshold is set with respect to $p_{\text{train}}(\x)$,
typically based on a fixed false positive rate \cite{yang2022}.

\begin{figure*}[t]
    \centering
    \includegraphics[width=.9\textwidth]{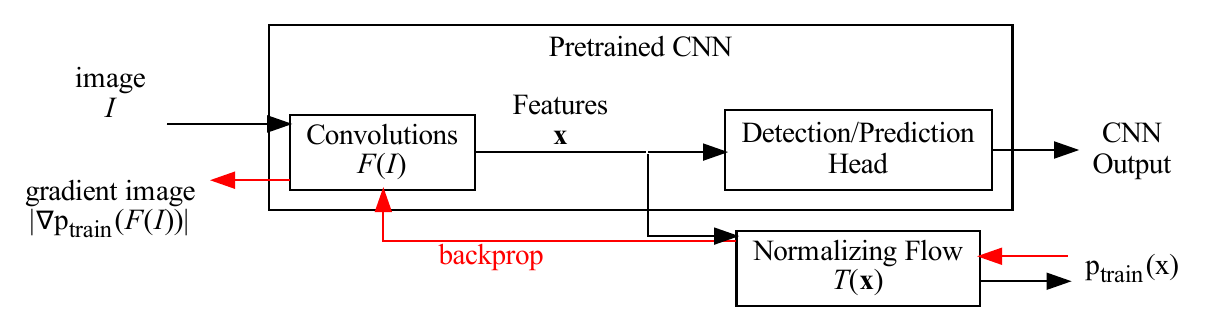}
    \caption{Illustration of how we apply NFs alongside a pretrained CNN for vision tasks.}
    \label{fig:nf_setup}
\end{figure*}

The primary difficulty in constructing an NF is ensuring invertibility, along with a tractable $|\det J_{T}(\x)|$.
One common way to solve these problems is \textit{affine coupling layers},
used in for example RealNVP\cite{RealNVP} and GLOW\cite{GLOW}.
Coupling layers divide their input $\x$ into two non-overlapping parts $\x_1, \x_2$.
An affine transformation is then performed on one part, while leaving the other unchanged:
\begin{equation*}
\begin{gathered}
    \x_1' = \a \cdot \x_1 + \b \enskip, \\
    \text{where} \ \a = \exp(F_1(\x_2)), \ \b = F_2(\x_2) \enskip.
\end{gathered}
\end{equation*}
Here we use $\cdot$ to denote element-wise multiplication.
$F_1$ and $F_2$ denote arbitrary functions, for example neural networks.
The output of the coupling layer is then computed by simply reconstructing $\x$ using the new $\x_1'$ and the unchanged $\x_2$.
Commonly, $\x_1$ and $\x_2$ are constructed by simply splitting $\x$ in half,
in which case the output is constructed by simply concatenating: $\x' = [\x_1' | \x_2]$ \cite{papamakarios}.
In this setting, the layers typically operate on alternating halves of $\x$,
otherwise one half would be left permanently unchanged.

Kirichenko \textit{et al.} \cite{kirichenko} observe that
applying NFs directly to images tends to work poorly for OOD detection,
They show that NFs seem to learn pixel-to-pixel correspondences,
rather than extracting any semantic information about the image distribution.
As a solution, they propose to apply an NF to features extracted from another pretrained neural network.

In our previous work \cite{scia23}, we adopted the approach from \cite{kirichenko},
and applied it to a robotic perception setting.
Instead of using a pretrained network only for feature extraction,
we applied an NF alongside a pretrained network for a vision task.
This approach allows a single network to perform a vision task,
while also being used as a feature extractor for the NF as illustrated in Fig. \ref{fig:nf_setup}.
In turn, this allows the NF to provide an OOD score for each image with negligible runtime overhead.
Using this NF construction,
we applied an evolutionary optimization procedure \cite{blackbox} to minimize $p_{\text{train}}(\x)$ by adjusting camera parameters
(backlight compensation, brightness, contrast, exposure, gain, saturation, and sharpness).
We implemented this optimization procedure in our robot, using YOLOv4\cite{YOLOv4} as the pretrained CNN.
By deploying our robot in a number of difficult visual scenarios,
we were able to show an increase in object detection performance using our optimization procedure
(average increase of 3.86 percentage units in F1 score).

\section{Our Approach}
Instead of computing a global OOD score,
we extend the NF approach from \cite{scia23} and \cite{kirichenko} by computing gradients to
extract pixel-level OOD information.
Formally, given an input image 
\begin{gather*}
    I = \begin{bmatrix}
        i_{11} & i_{12} & \ldots & i_{1W} \\
        i_{21} & i_{22} & \ldots & i_{2W} \\
        \vdots & \vdots & \ddots & \vdots \\
        i_{H1} & i_{H2} & \ldots & i_{HW}
    \end{bmatrix}, \ \
    I \in \mathbb{R}^{H \times W} \enskip, 
\end{gather*}
the OOD detection method in \cite{scia23,kirichenko} computes:
\begin{equation}
\begin{split} \label{eq:fwd_pass}
    \x &= F(I), \ \ F: \mathbb{R}^{H \times W} \mapsto \mathbb{R}^D \\
    p_{\text{train}}(\x) &= p_u(T(\x)) | \det J_T(\x) | \enskip,
\end{split}
\end{equation}
where F denotes the feature extraction layers from a pretrained neural network, and $T$ denotes the learned NF.

In order to gain pixel-level OOD information, we propose to use the gradient of Eq. (\ref{eq:fwd_pass}).
Specifically, we propose to use the absolute value of each element in the gradient.
With some notational abuse, we write
\begin{equation} \label{eq:grad}
\begin{split}
    | \nabla p_{\text{train}}(F(I)) | \in \mathbb{R}^{H \times W} \\
    | \nabla p_{\text{train}}(F(I)) |_{yx} = \Big|\dfrac{\delta p_{\text{train}}}{\delta i_{yx}}\Big| \enskip.
\end{split}
\end{equation}
In other words, we simply backpropagate through $p_{\text{train}}$ and $F$, as indicated by the red edges in Fig. \ref{fig:nf_setup}.
Here we use 2-dimensional grayscale images purely for notational convenience.
Our method extends directly to RGB images by averaging across the channel dimension.

Through initial testing
we have observed that image regions that closely match the distribution $p_{\text{train}}$ tend to have small gradient values
compared to regions that are OOD.
Intuitively, we reason that in-distribution data will likely lie close to a maximum of $p_{\text{train}}(\x)$.
Due to the continuity and differentiability of $p_{\text{train}}(\x)$,
these data points should exhibit small gradient values.
This intuition also serves as our rationale for using the absolute value of gradients,
as the direction of change is inconsequential.

Just as before, our goal is to construct an optimization procedure over camera parameters,
in order to allow our robot to adaptively overcome difficult visual scenarios.
In our robotics work, we mostly perform object detection tasks, for example object picking.
All object detection tasks involve some type of region proposal, whether it be bounding-boxes or image segmentations.
For such tasks, we propose to optimize:
\[
    \min_{\theta} \ \
    \frac{1}{|\text{ROI}|}\sum_{x, y \in \text{ROI}} | \nabla p_{\text{train}}(F(I(\theta))) |_{xy} \enskip,
\]
where ROI denotes the region of interest proposed by the object detector,
and $\theta$ denotes camera parameters.
We use $|\text{ROI}|$ to represent the number of pixels inside the region.
Here we also write $I(\theta)$ to highlight the dependence of the image on the camera parameters.
When the object detector fails to propose a suitable region of interest,
we fall back to computing the average over the entire absolute NF gradient image $|\nabla p_{\text{train}}(F(I(\theta)))|$.
We expect this fallback behaviour to perform similarly to optimizing $p_{\text{train}}$, as in \cite{scia23}.
Additionally, for multi-object detectors, we apply the sum over the union of all proposed regions:
\[
    \text{ROI} = \bigcup_{\text{ROI}_i} \text{ROI}_i \enskip.
\]

For other types of visual tasks, it may be necessary to construct other ways to select suitable regions to optimize over.

\section{Experiments and Results}
In this section, we begin by performing two separate experiments to empirically
establish evidence supporting our proposed optimization procedure.

First, we perform a conventional OOD detection experiment using the average absolute NF gradient as an OOD score:
\begin{equation} \label{eq:avg_abs_grad}
    \frac{1}{HW} \sum_{y=1}^H \sum_{x=1}^W \Big| \frac{\delta p_{\text{train}}}{\delta i_{yx}} \Big| \enskip.
\end{equation}
Showing that the absolute NF gradient can be used successfully for OOD detection
directly implies that it is useful for constructing an adaptive optimization procedure as in \cite{scia23}.
Additionally, showing success in OOD detection using the average absolute gradient over the entire image supports our
decision to use this as a fallback when our object detection fails to propose a region of interest.

Second, we investigate the relationship between object detector performance, and the absolute NF gradient.
Specifically, we look at the predicted bounding-boxes from two common object detectors
(YOLOv4\cite{YOLOv4}, Faster-RCNN\cite{FasterRCNN}),
and we show that bounding-boxes that are predicted correctly tend to contain smaller absolute NF gradient values
compared to incorrectly predicted bounding-boxes.
This experiment serves directly to support our decision to optimize camera parameters over the
predicted regions of interest from our object detector.

Finally, we conduct a real-world experiment using our robot, where we set up an object picking task in a cluttered scene.
We manually construct an exceedingly difficult visual scenario by making our lab as dark as possible,
while aiming a bright light at our robot.
Our proposed absolute NF gradient optimization procedure is then compared to \cite{scia23},
and simple auto-exposure and auto-whitebalance.

\subsection{NF Training}
In all of our experiments we use an NF consisting of 10 affine coupling layers, operating on alternating halves of the input.
Each coupling layer consists of two shared ReLU-activated linear layers followed by two separate prediction heads,
one for $\a$ and one for $\b$.
Our NF is trained for 200 epochs, using the Adam\cite{Adam} optimizer with a constant learning rate of $10^{-4}$.

\subsection{OOD Experiment}
We run the same conventional OOD detection experiment as in \cite{kirichenko},
in order to compare performance of our proposed average absolute NF gradient to that of using $p_{\text{train}}$ directly.
For feature extraction, we use an EfficientNet \cite{EfficientNet} model pretrained on ImageNet \cite{ImageNet}.
We perform OOD detection on the CIFAR-10 \cite{CIFAR10}, SVHN \cite{SVHN}, and CelebA \cite{CelebA} datasets.
An NF model is trained for each of these datasets, which results in the other two datasets being considered OOD.
First, we compute OOD detection performance using $p_{\text{train}}(F(I))$ as an OOD score directly, exactly like \cite{kirichenko}.
Then, we compute OOD detection performance using the average absolute NF gradient (Eq. \ref{eq:avg_abs_grad}).
We calculate an OOD threshold with a fixed 5\% false-positive rate.
The resulting true-positive rates are shown in Table \ref{tab:ood_px} and \ref{tab:ood_grad} respectively.

\begin{table}[b]
    \centering
    \caption{OOD detection true-positive rates using $p_{\text{train}}(F(I))$.} 
    \label{tab:ood_px}
    \begin{tabular}{|c|c|c|c|c|}
        \hline
        \multicolumn{2}{|c|}{} & \multicolumn{3}{|c|}{\textbf{OOD}} \\
        \cline{3-5}
        \multicolumn{2}{|c|}{} & \textbf{Cifar10} & \textbf{SVHN} & \textbf{CelebA} \\
        \hline
        \multirow{3}{6ex}{\textbf{Train}} & \textbf{Cifar10} & - & 0.4148 & 0.9953 \\
        \cline{2-5}
         & \textbf{SVHN} & 0.999 & - & 1.0 \\
        \cline{2-5}
         & \textbf{CelebA} & 1.0 & 1.0 & - \\
        \hline
    \end{tabular}
\end{table}

\begin{table}[b]
    \centering
    \caption{OOD detection true-positive rates using average absolute NF gradient.}
    \label{tab:ood_grad}
    \begin{tabular}{|c|c|c|c|c|}
        \hline
        \multicolumn{2}{|c|}{} & \multicolumn{3}{|c|}{\textbf{OOD}} \\
        \cline{3-5}
        \multicolumn{2}{|c|}{} & \textbf{Cifar10} & \textbf{SVHN} & \textbf{CelebA} \\
        \hline
        \multirow{3}{6ex}{\textbf{Train}} & \textbf{Cifar10} & - & 0.6688 & 0.9205 \\
        \cline{2-5}
         & \textbf{SVHN} & 0.972 & - & 0.9999 \\
        \cline{2-5}
         & \textbf{CelebA} & 0.9464 & 0.9747 & - \\
        \hline
    \end{tabular}
\end{table}

From Table \ref{tab:ood_grad}, it is evident that the NF gradient works well for OOD detection.
While we do observe a minor decrease in overall OOD detection performance,
our NF gradient is not far behind the plain NF approach, and even achieves improved performance on CIFAR10/SVHN.
We would especially like to stress that we do not aim to achieve state-of-the-art OOD detection performance,
but only highlight the usefulness of accessing this type of information at a pixel-level.

\subsection{Correlation with Model Performance}
In order to further establish our proposed method for adapting camera parameters,
we wish to investigate whether the absolute NF gradient correlates with object detection performance.
We perform an experiment using YOLOv4\cite{YOLOv4} and Faster-RCNN\cite{FasterRCNN}, both trained on the COCO\cite{COCO} dataset.
Both of these models output bounding-boxes corresponding to each predicted object.
In order to investigate the absolute NF gradient with respect to object detection performance,
we measure the average absolute NF gradients inside correctly classified bounding-boxes,
and compare it to incorrectly classified bounding-boxes.

If a predicted bounding-box overlaps at least 50\% (intersection-over-union) with a ground-truth box,
and they both have the same class label,
then we classify that predicted bounding-box as correct.
Otherwise it is classified as incorrect.
Similarly, if the model fails to predict a ground-truth box, we add that ground-truth box to our list of incorrect boxes.
These results are shown in Table \ref{tab:comp_coco}.

Table \ref{tab:comp_coco} shows average absolute NF
gradients inside bounding-boxes from the COCO validation dataset.
\begin{table}[t]
    \vspace{1em}
    \centering
    \centering
    \caption{Average gradients for COCO object detection.}
    \label{tab:comp_coco}
    \begin{tabular}{|c|c|c|}
        \hline
        & \textbf{Correct bounding-box} & \textbf{Incorrect bounding-box} \\
        \hline
        \textbf{YOLOv4} & 0.4447 & 0.5640 \\
        \hline
        \textbf{Faster-RCNN} & 0.7658 & 0.9510 \\
        \hline
    \end{tabular}
\end{table}
Since we operate on the absolute values of the NF gradients,
the resulting distributions are not Gaussian, and standard deviations appear meaningless.
Therefore, it is difficult to precisely gauge the nature of this correlation.
But what is evident is that there is a correlation,
as the correct bounding boxes have on average around 20\% lower absolute gradient values.

\subsection{Parameter Optimization}
\begin{figure}[b]
    \centering
    \includegraphics[width=.45\textwidth]{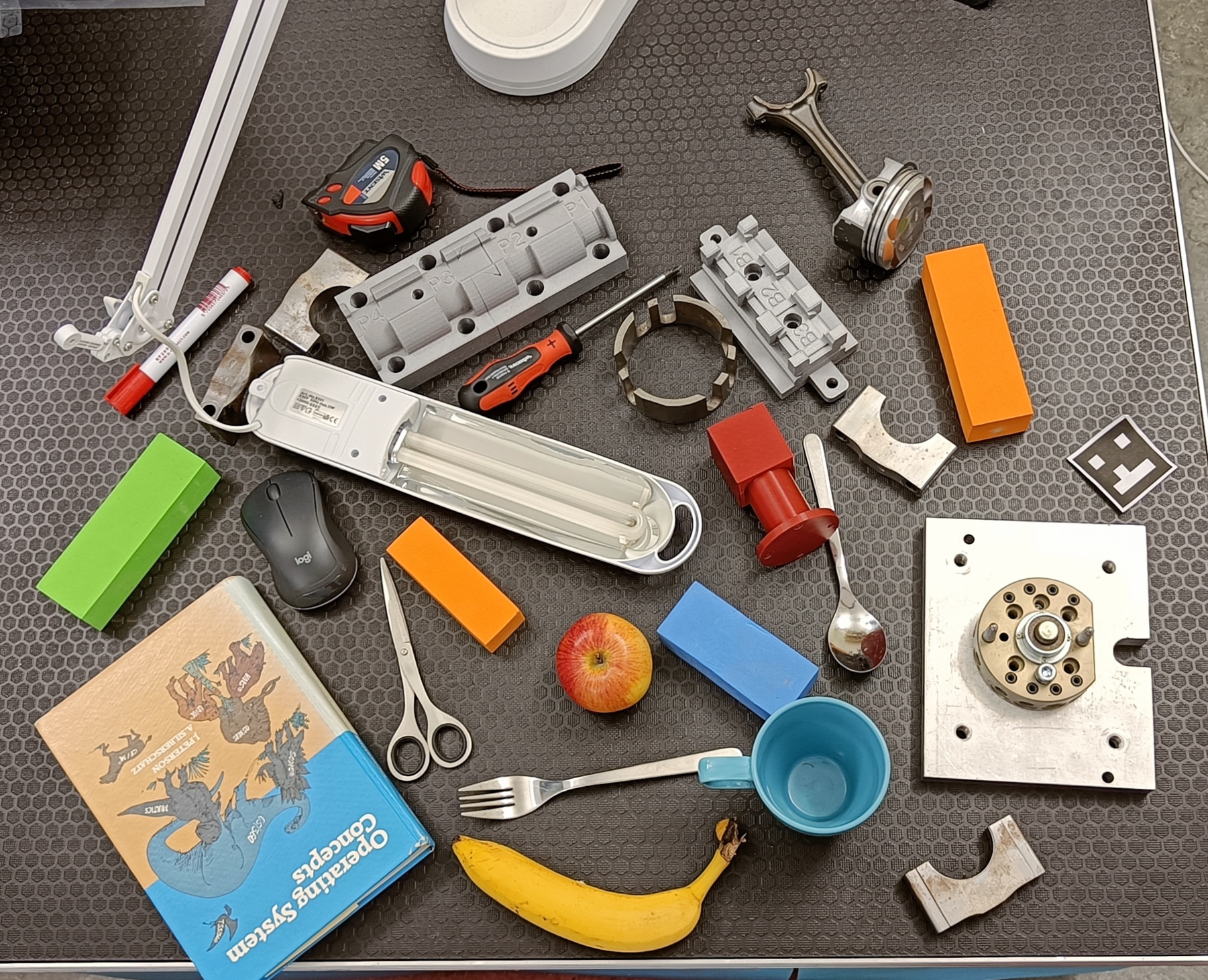}
    \caption{Cluttered scene used in parameter optimization experiment.}
    \label{fig:object_scene}
\end{figure}

\begin{figure*}[t]
    \vspace{1em}
    \centering
    \centering
    \parbox{.91\textwidth}{
        \centering
        \includegraphics[width=.45\textwidth]{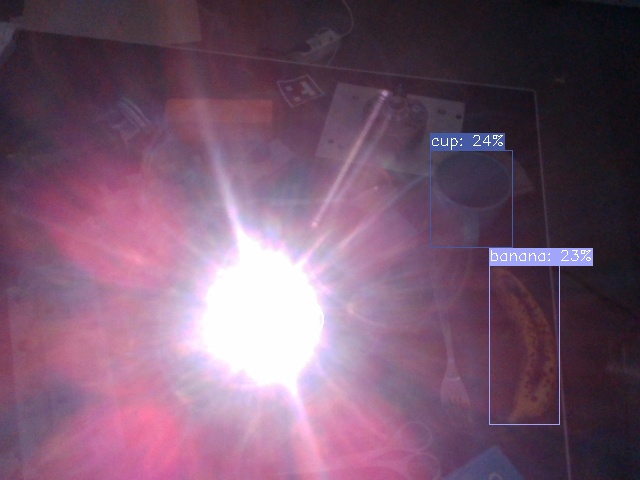}
        \includegraphics[width=.45\textwidth]{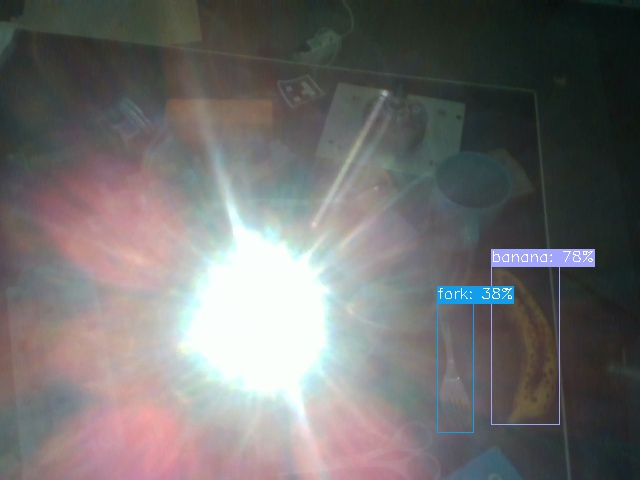} \\[1ex]
        \includegraphics[width=.45\textwidth]{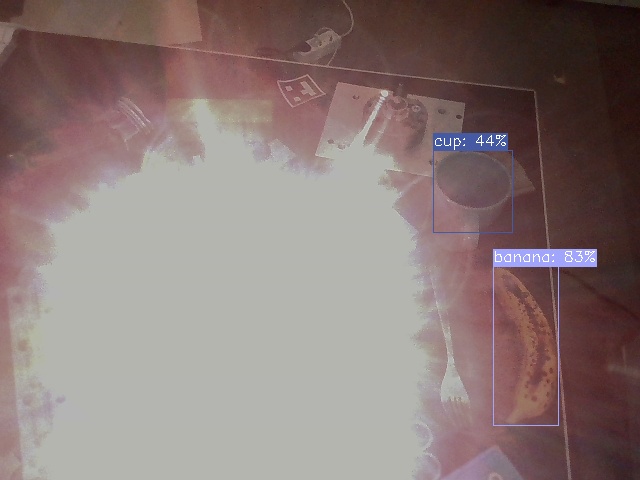}
        \includegraphics[width=.45\textwidth]{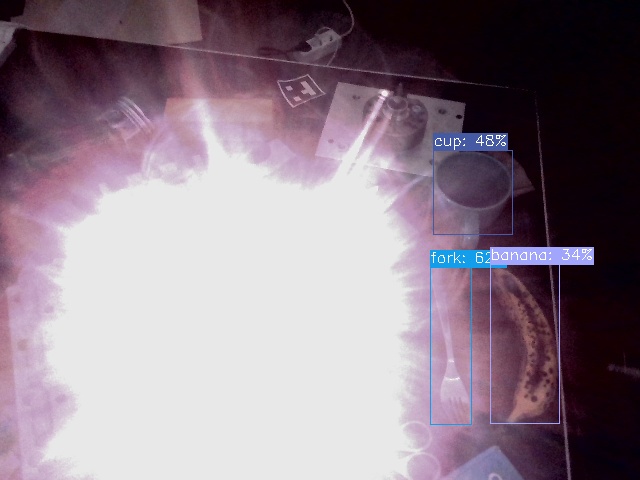}
    }
    \caption{
        Example images captured during experiments.
        Top left: default parameters.
        Top right: Default parameters with auto-exposure and auto-whitebalance.
        Bottom left: parameters optimized according to \cite{scia23}.
        Bottom right: parameters optimized using our NF gradient image.
    }
    \label{fig:example_images}
\end{figure*}

In order to show the benefits of our proposed absolute NF gradient for robotic perception,
we construct a table-top object picking task with difficult imaging conditions.
First we create a cluttered tabletop scene (Fig. \ref{fig:object_scene}).
Then, we make efforts to block out natural light to make the room as dark as possible,
and we shine a strong light facing straight up from the cluttered scene.
With all that in place, we position a UR-5 robot arm with a camera at its wrist (Realsense D435 RGB-D camera),
such that the camera is looking at our cluttered scene from above.
Our experiment then proceeds in an automated fashion as follows:
\begin{itemize}
    \item The arm is positioned such that the camera is looking down at the scene from a random position on a semi-sphere.
    \item Images are captured with:
        \begin{itemize}
            \item The camera's default image parameters.
            \item The camera's default image parameters, with auto-exposure and auto-whitebalance.
            \item Parameters optimized using $p_{\text{train}}(\x)$, as in \cite{scia23}.
            \item Parameters optimized using our proposed absolute NF gradient image.
        \end{itemize}
    \item For each image, a YOLOv4 model detects COCO objects in the scene.
\end{itemize}
Example images captured from one arm position can be seen in Fig. \ref{fig:example_images}.
In our scene we include the following COCO objects: apple, banana, book, cup, fork, mouse, scissors, and spoon
(Fig. \ref{fig:object_scene}).
We ensure that all COCO objects are visible to a human under some camera parameter settings.
Before starting our experiment, we compute the true position of each COCO object,
and convert these positions into the robot's coordinate frame.
For each trial in our experiment,
we compare the outputs from YOLOv4 to these positions in order to
determine whether an object was correctly classified at the right location.

Given that we use the same camera model as in \cite{scia23}, we choose to optimize the same parameters:
backlight compensation, brightness, contrast, exposure, gain, saturation, and sharpness.

For both parameter optimization procedures, we employ an evolutionary optimization algorithm with elitism selection,
population size=50, mutation rate=20\%, and a total of 200 iterations \cite{blackbox}.
All of these hyperparameters are the same as in \cite{scia23}.

We ran our experiment for a total of 83 trials from different views.
The number of correct predictions for each object class are shown in Table \ref{tab:results}.
Best results for each object class are highlighted in bold.

\begin{table}[t]
    \centering
    \centering
    \caption{
        Number of correct object detections.
        Default: Camera's default parameters.
        Auto: Camera's default parameters with auto-exposure and auto-whitebalance.
        NF: Parameters optimized w.r.t. $p_{\text{train}}(\x)$ as in \cite{scia23}.
        NF Gradient: Parameters optimized w.r.t. our proposed NF gradient image.
    }
    \label{tab:results}
    \begin{tabular}{|l|c|c|c|c|}
        \hline
         & \textbf{Default} & \textbf{Auto} & \textbf{NF} & \textbf{NF Gradient} \\
        \hline
        \textbf{apple} & 0 & 0 & 0 & 0 \\
        \hline
        \textbf{banana} & 1 & 8 & 6 & \textbf{16} \\
        \hline
        \textbf{book} & 5 & 2 & 6 & \textbf{11} \\
        \hline
        \textbf{cup} & 9 & 12 & 12 & \textbf{18} \\
        \hline
        \textbf{fork} & 24 & 19 & 34 & \textbf{45} \\
        \hline
        \textbf{mouse} & 0 & 0 & 0 & 0 \\
        \hline
        \textbf{scissors} & 2 & 0 & 4 & \textbf{10} \\
        \hline
        \textbf{spoon} & 4 & 1 & 9 & \textbf{14} \\
        \hline
        \hline
        \textbf{Total:} & 45 & 42 & 71 & \textbf{114} \\
        \hline
    \end{tabular}
\end{table}

It is immediately clear from Fig. \ref{fig:example_images} that these imaging conditions are very difficult for
an object detector to operate in, which explains the low detection numbers in Table \ref{tab:results}.
Table \ref{tab:results} shows that optimizing camera parameters with respect to our proposed NF gradient image
yields by far the best object detection performance for every object class we tested.
In fact, our method achieves an overall success rate that is 60\% better than the second best method.
For the individual classes, our method beats the second best by at least 32\% (\textit{fork}), and up to 150\% (\textit{scissors}).
We disregard the \textit{apple} and \textit{mouse} object classes, since they were not detected by any method.

While we ensured that all objects were visible under some parameter settings,
we never found a single parameter setting that made all objects clearly visible.
Most likely, this explains why neither method could successfully detect the \textit{apple} and \textit{mouse}.
Since they were placed close to the light source, they were typically more difficult to discern.
This further highlights the importance of local OOD information compared to a single global OOD score.
By computing a local OOD score,
our NF gradient method allows the perception pipeline to focus only on the regions that matter, and make sure those are visible.

We would like to note that it is likely possible to further improve the object detection performance.
In our experiment, camera parameters were jointly optimized for all detected objects.
However, we believe that individual optimization for each object region might yield even better performance.

\section{Conclusions and Future Work}
In this work, we introduce a technique that employs gradients from a
Normalizing Flow (NF) to obtain pixel-level Out-of-Distribution (OOD) information for an image.
Experimental results validate the efficacy of our proposed NF gradient for general OOD detection. 
Moreover, this approach significantly surpasses our prior methods for adapting to challenging imaging scenarios. 
Consequently, our robotic perception framework demonstrates enhanced object detection capabilities,
even in the face of adverse lighting conditions.

We will make our code publicly available at
\href{https://github.com/SimonKLind/NF-Gradients}{https://github.com/SimonKLind/NF-Gradients}.
Additionally, we will compile the images collected from our experiment into a dataset,
which we will also make publicly available for future research.

A notable limitation of our current method is its dependency on the camera to
sample parameter configurations during the optimization procedure. 
The process of acquiring multiple image samples is inherently time-consuming,
resulting in a runtime that is slower than desired,
which may prohibit its use in some robot applications.
To broaden the applicability of our techniques across diverse robotic tasks,
we hope to improve the runtime by removing the dependency on the camera for sampling.

\section*{ACKNOWLEDGMENTS}
This research is funded by the Excellence Center at Linköping-Lund in Information Technology (ELLIIT),
and the Wallenberg AI, Autonomous Systems and Software Program (WASP).
Computations for this publication were enabled by the supercomputing resource Berzelius provided by the National Supercomputer Centre at Linköping University and the Knut and Alice Wallenberg foundation.

\bibliographystyle{IEEEtran}
\bibliography{IEEEabrv,paper}

\end{document}